\documentclass[10pt,twocolumn,letterpaper]{article}

\usepackage{cvpr}              

\usepackage{multirow}
\usepackage{array}
\usepackage{algorithm}
\usepackage{algpseudocode}
\usepackage[shortlabels]{enumitem}
\usepackage{stfloats}
\usepackage{cuted}

\definecolor{cvprblue}{rgb}{0.21,0.49,0.74}
\usepackage[pagebackref,breaklinks,colorlinks,allcolors=cvprblue]{hyperref}

\def\paperID{43306} 
\def\confName{CVPR}
\def\confYear{2026}

\title{SimRecon: SimReady Compositional Scene Reconstruction from Real Videos}

\begin{document}


\author{
    Chong Xia\textsuperscript{1,2,\raisebox{2pt}{$\star$},*}\quad
    Kai Zhu\textsuperscript{1,*}\quad
    Zizhuo Wang\textsuperscript{1}\quad
    Fangfu Liu\textsuperscript{1}\quad
    Zhizheng Zhang\textsuperscript{2}\quad
    Yueqi Duan\textsuperscript{1,\dag}
    \vspace{1mm}\\
    \textsuperscript{1}Tsinghua University \quad
    \textsuperscript{2}Galbot
    \vspace{1mm}\\
    Project Page: \url{https://xiac20.github.io/SimRecon/}
}

\maketitle

\begingroup
\renewcommand\thefootnote{} 
\footnotetext{\textsuperscript{*} Equal contribution. \quad \textsuperscript{\dag} Corresponding author.} 
\footnotetext{\textsuperscript{$\star$} Work done during an internship at Galbot.} 
\endgroup

\begin{strip}
\vspace{-20mm}
    \centering
    \includegraphics[width=1.0\linewidth]{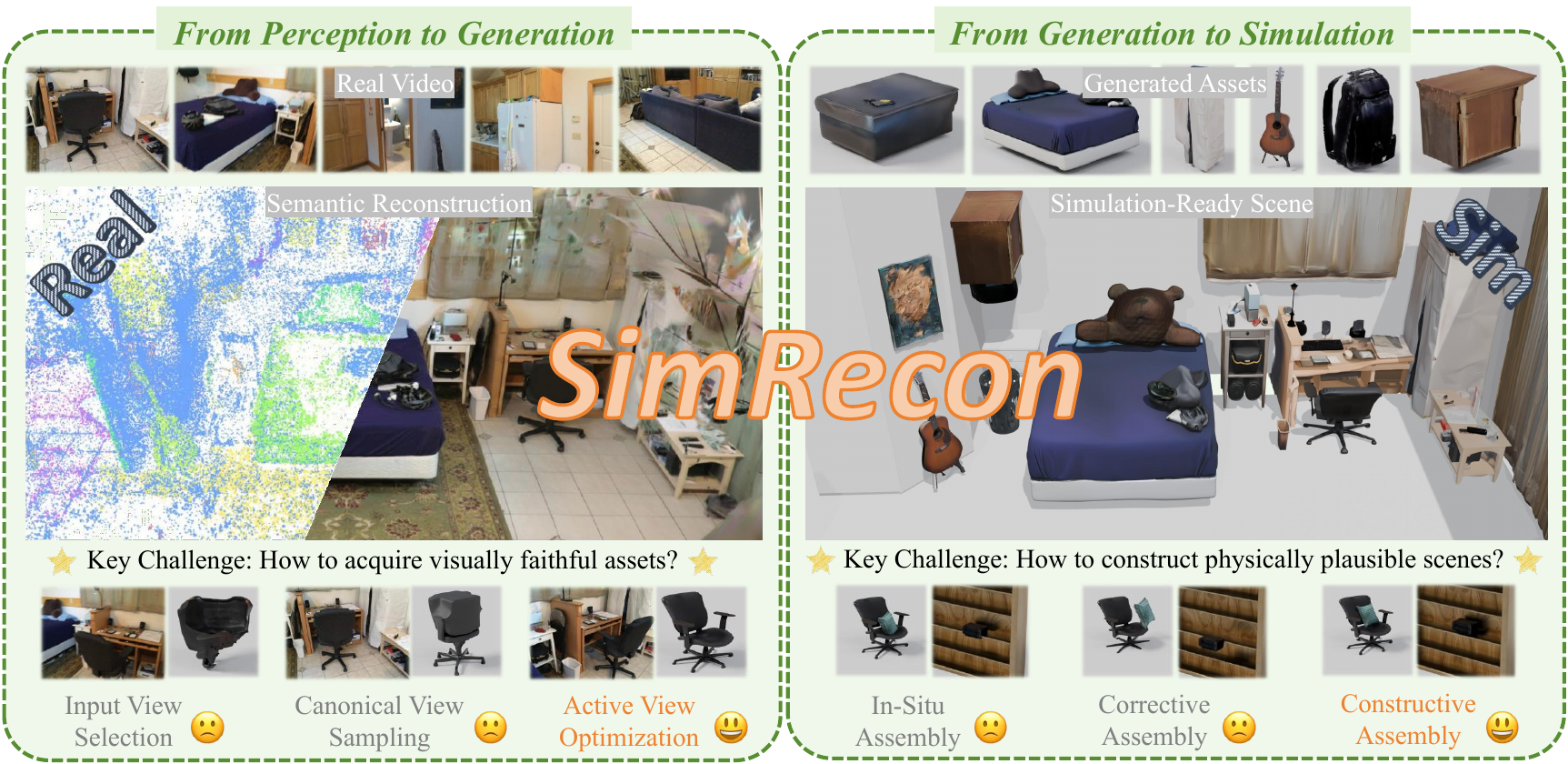}
   \captionof{figure}{\textbf{We propose SimRecon, a framework for reconstructing simulation-ready, compositional 3D scenes from real-world videos.} Our method introduces a ``Perception-Generation-Simulation'' pipeline which transforms cluttered input videos into physically assembled scenes. To ensure visually faithful generated assets and physically plausible final scenes, we propose two bridging strategies: Active View Optimization to acquire optimal generation conditions, and Constructive Assembly to follow a native building principle.}
    \label{teaser}
\end{strip}



\begin{abstract}
Compositional scene reconstruction seeks to create object-centric representations rather than holistic scenes from real-world videos, which is natively applicable for simulation and interaction. Conventional compositional reconstruction approaches primarily emphasize on visual appearance and show limited generalization ability to real-world scenarios. In this paper, we propose SimRecon, a framework that realizes a ``Perception-Generation-Simulation" pipeline towards cluttered scene reconstruction, which first conducts scene-level semantic reconstruction from video input, then performs single-object generation, and finally assembles these assets in the simulator. However, naively combining these three stages leads to visual infidelity of generated assets and physical implausibility of the final scene, a problem particularly severe for complex scenes. Thus, we further propose two bridging modules between the three stages to address this problem. To be specific, for the transition from Perception to Generation, critical for visual fidelity, we introduce Active Viewpoint Optimization, which actively searches in 3D space to acquire optimal projected images as conditions for single-object completion. Moreover, for the transition from Generation to Simulation, essential for physical plausibility, we propose a Scene Graph Synthesizer, which guides the construction from scratch in 3D simulators, mirroring the native, constructive principle of the real world. Extensive experiments on the ScanNet dataset validate our method's superior performance over previous state-of-the-art approaches.

\end{abstract}

\section{Introduction}
3D scene reconstruction from multi-view images is a long-standing challenge in computer vision. Recent advances in neural representations~\cite{kerbl3Dgaussians, mildenhall2020nerf} have enabled significant progress in 3D geometry reconstruction~\cite{oechsle2021unisurf, wang2021neus, yariv2021volume} and novel view rendering~\cite{barron2022mip, gao2024cat3d, wu2023reconfusion, yang2023freenerf}. However, these methods represent the scene holistically: although they achieve impressive visual fidelity, they remain fundamentally unsuitable for simulation and interaction since they lack complete object geometry and well-defined object boundaries. Concurrently, contemporary studies have focused on creating 3D indoor simulators by manually placing assets within simulated environments~\cite{ge2024behavior, khanna2024habitat, li2021igibson, puig2023habitat}, by using specialized capture hardware during scanning~\cite{baruch2021arkitscenes, straub2019replica, yeshwanthliu2023scannetpp, yu2025metascenes} with extensive manual annotation, or by employing procedural generation via rule-based~\cite{deitke2022, paschalidou2021atiss, raistrick2024infinigen} or learned layout generative models~\cite{tang2023diffuscene, yang2024physcene, yang2024holodeck}. These datasets have significantly advanced Embodied AI research, particularly in embodied reasoning~\cite{das2018embodied, majumdar2024openeqa, shridhar2020alfred}, navigation~\cite{hong2021vln, jiang2024autonomous, jiang2024scaling, szot2021habitat}, and manipulation~\cite{ge2024behavior, huang2023embodied, khanna2024habitat}. Nonetheless, these scene creation methods still depend on well-reconstructed scan data with extensive manual engagement, and suffer from artificial layouts that diverge from the real world.

A new branch of work has begun to explore compositional 3D reconstruction from only multi-view images in the wild~\cite{li2023rico, wu2023objsdfplus, ni2025decompositional, yang2025instascene}, but several key limitations in these approaches hinder this goal. First, these methods often rely on heuristic view selection from the input images or 3D representation for single-object generation, which struggles to produce complete and plausible geometry for small, large or occluded objects. Second, their final result is still a visual representation rather than a simulation-ready scene, leading to a ``real-to-sim" gap manifested as physical implausibility. Third, they often rely on specially designed methods for semantic reconstruction and object generation, which are tightly coupled to their own pipeline and cannot easily leverage advanced approaches in these areas.

In this paper, we propose SimRecon, a framework that realizes a ``Perception-Generation-Simulation" pipeline with a unified object-centric spatial representation, aiming at transforming the clutter video input to a simulation-ready compositional 3D scene. Our framework starts with semantic reconstruction from video input to restore 3D scene and differentiate individual objects, then conducts single-object generation to complete each instance, and finally assembles these assets within a physical simulator. The primary challenges are the visual infidelity of generated assets and physically implausibility of the final constructed scenes, which derive from the connection parts from the three stages. Building upon this observation, we mainly focus on designing bridging modules to address these bottlenecks: achieving complete geometry and appearance for individual objects, and ensuring their physically plausible placement. The bridging module design paradigm also endows our framework with inherent extensibility.

Specifically, to bridge the gap from perception to generation, which requires converting unstructured and cluttered 3D geometric representations into effective image conditions for generation models, we introduce Active Viewpoint Optimization, which intelligently searches for optimal views in the 3D scene with maximized information gain as the best view condition. This method moves beyond heuristic view selection, which often yields occluded views in complex scenes and leads to deformed generated assets. Moreover, to ensure plausible scene construction in the simulator, we introduce Scene Graph Synthesizer, which progressively extracts a global scene graph from multiple incomplete observations. This scene graph mainly models the supportive and attached relations among objects, which serves as the native constructive guideline for the following hierarchical physical assembly to ensure physical plausibility. Extensive experiments on the ScanNet dataset demonstrate the superiority of our approach over state-of-the-art methods in terms of reconstruction fidelity for complex scenes and physical plausibility in the simulator.

\section{Related Work}

\paragraph{3D Indoor Scene Simulators.} Recent efforts have focused on creating 3D indoor scene simulators for embodied tasks, which are mainly categorized into three types based on their scene construction methods: hand-crafted, generation-based, and scan-based. Hand-crafted methods~\cite{ge2024behavior, khanna2024habitat, li2021igibson, puig2023habitat} manually design scene layouts and place assets within simulated environments, requiring extensive manual annotation. With the development of VLMs~\cite{achiam2023gpt, lu2024deepseek, wang2024qwen2} and diffusion models~\cite{ho2020denoising}, many generative works employ procedural scene generation with rule-based commensense priors~\cite{deitke2022, paschalidou2021atiss, raistrick2024infinigen} or learned layout priors~\cite{tang2023diffuscene, yang2024physcene, yang2024holodeck}. However, both hand-crafted and generative methods often result in layouts that are overly simplistic and deviate from real-world complexity. Scan-based approaches, conversely, offer superior realism and authenticity by leveraging data captured from real environments. However, these scanning methods rely on specialized capture devices to acquire 3D point clouds or meshes and still require extensive manual annotation~\cite{hua2016scenenn, sun2018pix3d, dai2017scannet}, even with semi-automated post-processing~\cite{avetisyan2019scan2cad, dai2024acdc, yu2025metascenes}. Recent approaches~\cite{mu2025robotwin, li2024robogsim, wang2025embodiedgen} have begun to explore fully automated reconstruction of real table-top or specific scenes from a single image, often leveraging segmentation foundation models~\cite{kirillov2023segment, ravi2024sam, ren2024grounded} and 3D asset generation models~\cite{tang2024lgm, voleti2024sv3d, xu2024instantmesh, zhang2024clay}. Furthermore, in this paper, we aim to establish a fully automated pipeline for scene-level, simulation-ready reconstruction from raw video input, unlocking the potential to generate diverse simulation environments from arbitrary videos.

\begin{figure*}[t]
    \centering
    \includegraphics[width=\linewidth]{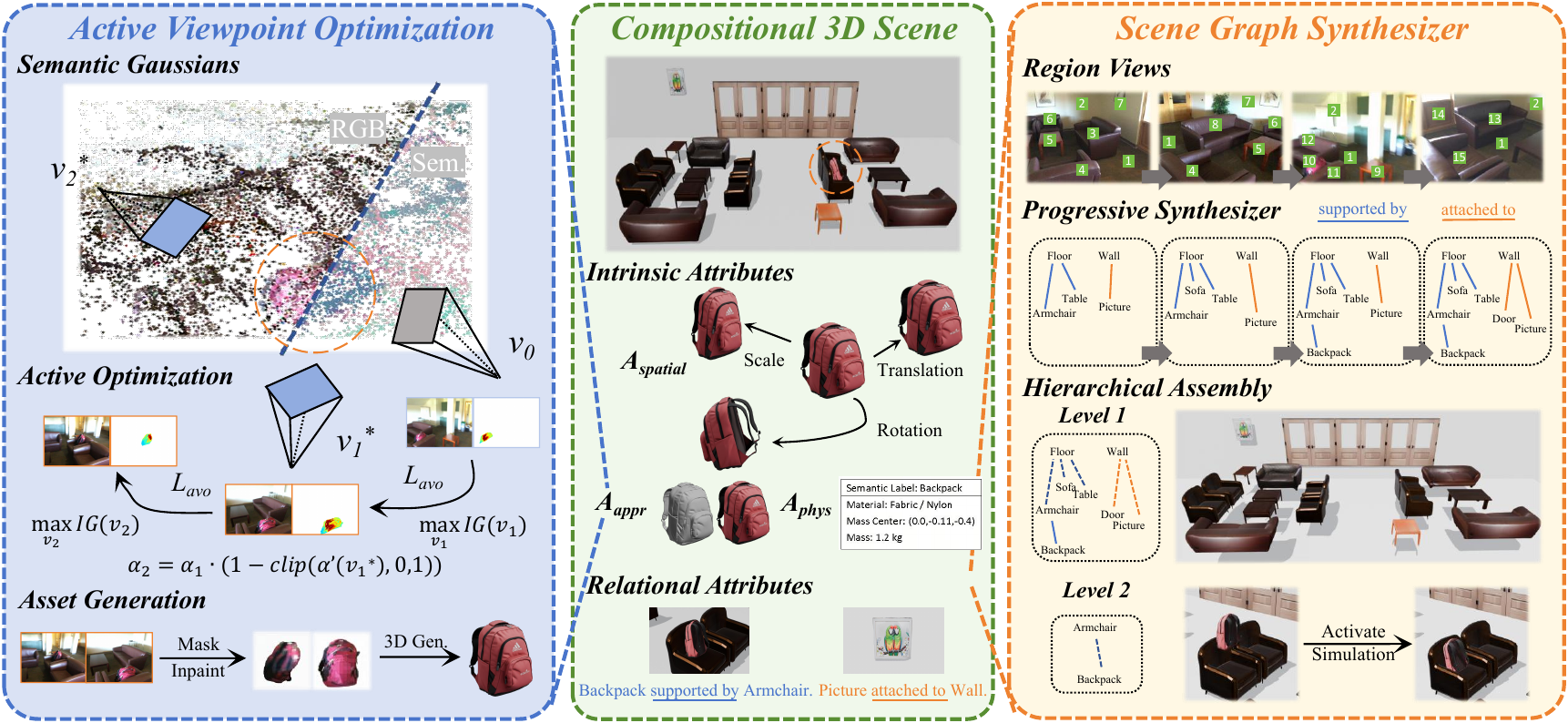}
    \caption{\textbf{The overall framework of our approach SimRecon.} We propose a ``Perception-Generation-Simulation" pipeline with object-centric scene representations towards compositional 3D scene reconstruction from cluttered video input. In this figure, we provide illustrative visualizations using the backpack as the example to introduce our two core modules: Active Viewpoint Optimization (AVO) and Scene Graph Synthesizer (SGS). There, we visualize a semantic-level graph for clarity, while our framework operates at the instance-level.}
    \label{fig:pipeline}
\end{figure*}

\paragraph{Compositional 3D Reconstruction.} Previous scene reconstruction approaches~\cite{kerbl20233d, mildenhall2021nerf, schoenberger2016mvs} usually model the entire scene as a holistic representation, whereas recent works have begun to focus on compositional 3D reconstruction methods~\cite{yao2025cast, ardelean2025gen3dsr, meng2025scenegen, xia2025drawer, huang2025midi, ni2025decompositional, yang2025instascene} for interactive scene generation and downstream embodied tasks. Early methods mainly focus on the simplified single-view scenarios, leveraging either a multi-stage pipeline~\cite{yao2025cast, ardelean2025gen3dsr} or an end-to-end generation paradigm~\cite{meng2025scenegen, huang2025midi}. Recent work DPRecon~\cite{ni2025decompositional} proposes a scene-level reconstruction pipeline, but its reliance on SDF~\cite{park2019deepsdf} and SDS~\cite{poole2022dreamfusion} with well-segmented input makes it time-consuming and hard to generalize to real scenarios. InstaScene~\cite{yang2025instascene} further leverages 3D semantic reconstruction to segment instances and specialized generation model to complete objects, but struggles with real scenes with complex objects and mainly focuses on visual appearance rather than simulation-ready scenes. In contrast, our framework robustly handles real complex scenes with fine-grained complete geometry for each object and finally constructs the corresponding simulation-ready scene within the physical simulator.


\paragraph{3D Scene Graphs.} 3D scene graph is a graph structure where nodes represent objects or areas, and edges encode pairwise relationships between them, such as spatial or functional connections. Traditional methods typically learn such graphs using Graph Neural Networks (GNNs) with 3D point clouds as input~\cite{Armeni2019_3dsg, Wald2020_3dssg, wu2021scenegraphfusion, wu2023incremental, hughes2022hydra}. However, with the advent of LLMs~\cite{achiam2023gpt, liu2024deepseek} and VLMs~\cite{lu2024deepseek,wang2024qwen2}, scene graphs recently can be inferred more easily through procedural queries. Nowadays, the 3D scene graph often serves as a concise scene representation, acting as a fundamental structure for scene understanding and other downstream tasks. For example, OpenIN~\cite{tang2025openin} builds hierarchical open-vocabulary 3D scene graphs for robot navigation, while ScenePainter~\cite{xia2025scenepainter} utilizes learnable textual token graphs for 3D scene outpainting. In this work, we aim to build a scene graph in a progressive paradigm to model the supportive and attached relations among objects, serving as the guideline for the following construction within the simulator.

\section{Approach}
\label{method}
In this section, we present our method, SimRecon, which realizes a ``Perception-Generation-Simulation" pipeline for compositional 3D reconstruction. At first, we detail our object-centric scene representation and overall architecture in Section~\ref{sec:representation}. Next, in Section~\ref{sec:avo}, we introduce Active Viewpoint Optimization (AVO), an approach designed to extract maximally informative projection views in 3D space for each object, even robust under heavy occlusion in complex scenes. Furthermore, in Section~\ref{sec:sgg}, we present Scene Graph Synthesizer (SGS), a method to infer the global scene graph in an online paradigm to guide the final hierarchical physical assembly. The overall framework of SimRecon is illustrated in Figure~\ref{fig:pipeline}.


\subsection{Object-Centric Scene Representation}
\label{sec:representation}

\paragraph{Compositional Scene Primitives.}
Conventional holistic approaches, exemplified by 3D Gaussian Splatting~\cite{kerbl3Dgaussians}, represent a scene $\mathcal{S}_{\text{holistic}}$ as a vast collection of low-level rendering primitives, $\{g_i\}_{i=1}^N$. This representation is non-structural, lacking explicit object boundaries or semantics, thus inherently unsuitable for physical interaction or semantic reasoning. In contrast, our compositional framework defines the scene $\mathcal{S}_{\text{comp}}$ as a structured set of $L$ discrete, high-level object primitives $o_i$, which serve as the fundamental building blocks for the scene:
\begin{equation}
\label{eq:scene_comp}
\mathcal{S}_{\text{comp}} = \{o_1, o_2, \ldots, o_L\}.
\end{equation}
Each object primitive $o_i$ is a comprehensive entity defined by two categories of attributes: intrinsic attributes $\mathcal{A}_{\text{int}}$ and relational attributes $\mathcal{A}_{\text{rel}}$.

\paragraph{Intrinsic Attributes.}
The intrinsic attributes define the object $o_i$ in isolation, independent of its surrounding context. We formally represent this as a tuple mainly comprising three primary dimensions:
\begin{equation}
\label{eq:intrinsic_attr}
\mathcal{A}_{\text{int}, i} = (A_{\text{spatial}, i}, A_{\text{appr}, i}, A_{\text{phys}, i}).
\end{equation}
Here, $A_{\text{spatial}, i}$ denotes the spatial attributes, including its scale $s_i \in \mathbb{R}^3$, its rotation $R_i \in SO(3)$ and translation $t_i \in \mathbb{R}^3$, which together form the 6-DoF pose $T_i \in SE(3)$. $A_{\text{appr}, i}$ represents the appearance attributes, defined by a complete geometric mesh $\mathcal{M}_i$ with its corresponding PBR textures $\mathcal{T}_i$. Finally, $A_{\text{phys}, i}$ comprises physical attributes essential for simulation, including its semantic label $l_i$, material $\textit{mat}_i$, center of mass $c_i$, and mass $m_i$.

\paragraph{Relational Attributes.}
The relational attributes $\mathcal{A}_{\text{rel}}$ define the object's role and context within the scene by encoding supportive, spatial, and functional semantic relationships with other objects. These explicit interactions are organized into a structured Scene Graph $\mathcal{G} = (\mathcal{S}_{\text{comp}}, \mathcal{E})$, where $\mathcal{E}$ is the set of edges $e_{ij} = (o_i, r_{ij}, o_j)$ representing a relation $r_{ij}$ between two object primitives.

\paragraph{Overall Architecture.}
In our pipeline, these attributes are progressively populated, transforming raw image observations into simulation-ready entities. The initial semantic reconstruction stage provides the foundational set of attributes $\{s_i, T_i, l_i\}$ for each segmented object. The 3D asset generation stage, conditioned on actively optimized image projections, then completes the geometry $\mathcal{M}_i$ and appearance $\mathcal{T}_i$ and allows for the inference of the remaining physical attributes $\{\textit{mat}_i, c_i, m_i\}$. Finally, the scene graph $\mathcal{G}$ is constructed by our online graph merging method, where its supportive and attached relations guides the hierarchical scene construction within simulators, ensuring a physically stable and plausible 3D scene.

\subsection{Active Viewpoint Optimization}
\label{sec:avo}

\paragraph{View Projection as a Bottleneck.}
Images serve as a general-purpose and powerful condition for 3D generative models. However, the quality of these views, particularly in the presence of severe occlusion or partial observations, drastically impacts the fidelity of the generated asset. Conventional methods often resort to heuristic strategies, such as using the original input views or sampling canonical surrounding viewpoints. These static approaches often fail to sufficiently capture complete and informative observations of the object, often yielding low-quality, uninformative, or redundant views that lead to deformed assets, especially for complex scenes. To overcome this, we propose Active Viewpoint Optimization (AVO), a framework that actively optimizes for most informative viewpoints for each object.
\paragraph{Information Theory Formulation.}
We model the optimal view projection problem as an information gaining task in information theory, where the goal is to optimize a viewpoint $v$ that maximizes the information gain about the object's complete reconstructed geometry $X_i$, from the initial viewpoint $v_0$. The information gain is defined as the reduction in information entropy $H$ with a new viewpoint $v$:
\begin{equation}
\label{eq:ig_base}
IG(v) = H(X|v_0) - H(X|v).
\end{equation}
Considering directly computing this entropy is intractable, we propose a practical and differentiable proxy for $-H(X|v)$ based on the alpha-blending process inherent in 3D Gaussian Splatting rendering. Intuitively, a viewpoint that yields a rendering with high accumulated opacity signifies a more solid and informative observation, thus corresponding to higher negative entropy. Let $\alpha(p, v)$ denote the accumulated opacity rendered along the ray passing through pixel $p$ from viewpoint $v$, calculated using the standard volumetric rendering equation:
\begin{equation}
\label{eq:alpha_rendering}
\alpha(p, v) = \sum_{i \in \mathcal{N}_p} \alpha_i \prod_{j=1}^{i-1} (1-\alpha_j)
\end{equation}
where $\mathcal{N}_p$ are the Gaussians intersected by the ray for pixel $p$, ordered by depth, and $\alpha_i$ is the intrinsic opacity of the $i$-th Gaussian along the ray. We define our total information proxy $A(v)$ as the sum over pixels corresponding to the object of this rendered opacity map:
\begin{equation}
\label{eq:accumulated_opacity_integral}
A(v) = \sum_{p \in \mathcal{P}_{\text{obj}}(v)} \alpha(p, v)
\end{equation}
Maximizing this total accumulated opacity $A(v)$ serves as a differentiable surrogate for maximizing the information gain $IG(v)$, thus the final objective is:
\begin{equation}
\label{eq:ig_final_revised}
\max_{v} IG(v) = \max_{v} A(v) = \max_{v} \sum_{p \in \mathcal{P}_{\text{obj}}(v)} \alpha(p, v).
\end{equation}
This formulation directly leverages the differentiability of the Gaussian Splatting rendering pipeline for efficient gradient-based optimization.

\paragraph{Single View Optimization with Constraints.}
Our first objective is to find the single optimal viewpoint $v^*$ by maximizing the information gain proxy $A(v)$ defined in Eq.~\ref{eq:accumulated_opacity_integral}. We parameterize the view pose $T_v$ of viewpoint $v$ (using a quaternion $q$ for rotation and position $t$) and initialize the parameters from one input view $v_0$ that captures the target object. The the optimization loss $L_{IG}$ is defined as the negative of the information gain:
\begin{equation}
\label{eq:loss_avo_revised}
L_{IG}(v) = -A(v) = - \sum_{p \in \mathcal{P}_{\text{obj}}(v)} \alpha(p, v).
\end{equation}
Since standard 3DGS rendering is non-differentiable for camera parameters $T_v$, we enable their optimization by applying the relative camera transformation to the differentiable Gaussian parameters $\mathcal{G}$ instead. 

Furthermore, to prevent extreme cases, such as the viewpoint collapsing too close to the object surface, we introduce a depth regularization term $L_{depth}$. This regularizer encourages the rendered depth $D(p, v)$ at each object pixel $p \in \mathcal{P}_{\text{obj}}(v)$ to remain close to a target depth $d_{\text{target}}(s_i)$, which is determined proportionally to the object's size $s_i$. We formulate this using an averaged quadratic penalty:
\begin{equation}
\label{eq:loss_depth_L2}
L_{depth}(v) = \frac{\lambda_{\text{depth}}}{|\mathcal{P}_{\text{obj}}(v)|} \sum_{p \in \mathcal{P}_{\text{obj}}(v)} (D(p, v) - d_{\text{target}}(s_i))^2.
\end{equation}
Here, $\mathcal{P}_{\text{obj}}(v)$ are the pixels corresponding to the object rendered from view $v$. The full optimization objective is thus:
\begin{equation}
\label{eq:loss_single_final_L2}
L_{AVO}(v) = L_{IG}(v) + L_{depth}(v).
\end{equation}
The optimization then proceeds by iteratively updating $T_v$ based on the gradient signal derived from the Gaussian rendering parameters.

\paragraph{Iterative Viewpoint Expansion.}
To generate a set of $K$ informative views, we employ an iterative optimization strategy. At each iteration $k$, we seek the viewpoint $v_k^*$ that maximizes information gain based on the currently remaining potential information, represented by effective opacities $\alpha_i^{(k-1)}$ (initially $\alpha_i^{(0)}$). The viewpoint $v_k^*$ is found by minimizing the single-view loss $L_{AVO}^{(k)}(v)$, which computes accumulated opacity using effective $\alpha_i^{(k-1)}$:
\begin{equation}
v_k^* = \arg\min_{v} \left( - \sum_{p \in \mathcal{P}_{\text{obj}}(v)} \alpha(p, v | \{\alpha_i^{(k-1)}\}) + L_{depth}(v) \right).
\end{equation}
After finding $v_k^*$, we update the effective opacities via multiplicative decay, reducing $\alpha_i^{(k-1)}$ based on its rendered contribution $\alpha'_{i}(v_k^*)$ from the selected view:
\begin{equation}
\alpha_i^{(k)} = \alpha_i^{(k-1)} \cdot (1 - \text{clip}(\alpha'_{i}(v_k^*), 0, 1)).
\end{equation}
This decay ensures subsequent iterations naturally focus on less observed regions. The process repeats until $K$ views are generated or a coverage threshold is met (e.g., remaining $\sum \alpha_i^{(k)} < \eta \sum \alpha_i^{(0)}$). Finally, for each $v_k^*$, we render the object appearance, inpaint occlusions, and provide these complete views as conditions to the generative model.

\subsection{Scene Graph Synthesizer}
\label{sec:sgg}

\paragraph{Scene Graph as Physical Scaffolding.}
While the previous stage provides visually complete object assets, assembling them correctly within a simulator is also challenging. Direct in-situ placement based on initial positions or corrective post-processing placement often leads to physically implausible configurations like floating objects or penetrations. Therefore, we propose a constructive placement method to ensure the physical plausibility at all times, which builds on the understanding of physical interdependencies among objects. To achieve this, we construct a scene graph $\mathcal{G} = (\mathcal{N}, \mathcal{E})$ which explicitly encodes fundamental physical support and attachment relationships. However, inferring such a graph directly for an entire cluttered scene is challenging due to severe occlusions and the complexity of global reasoning. Therefore, we adopt a progressive approach, synthesizing the global graph incrementally from multiple local observations.

\paragraph{Region-based Scene Graph Inference.}
To implement this progressive synthesis, we first partition the set of object instances $\mathcal{S}_{\text{comp}}$ into $K$ spatial regions $\mathcal{C} = \{\mathcal{C}_1, \ldots, \mathcal{C}_K\}$ via DBSCAN~\cite{ester1996density} clustering on the object centroids $\{c_i\}_{i=1}^L$. Objects not assigned to any cluster are subsequently assigned to the spatially nearest cluster. For each region $\mathcal{C}_k$, an optimal observation viewpoint $v_k^*$ is obtained by adapting the Active Viewpoint Optimization objective to maximize information gain across all objects within $\mathcal{C}_k$. A projection image $I_k$ is rendered from $v_k^*$, annotated with the corresponding instance IDs for visible objects. This image $I_k$, along with the list of visible instance IDs, is fed to a Vision-Language Model (VLM) via a structured prompt to request ``(Child ID, Relation, Parent ID)'' triplets describing direct physical support (``supported\_by'') and attachment (``attached\_to'') relationships. Floor and wall entities are treated as initial nodes in this graph structure and serve as the physical foundation for other objects within the scene. This yields a local subgraph $\mathcal{G}_k = (\mathcal{N}_k, \mathcal{E}_k)$ per region.


\paragraph{Online Scene Graph Merging.}
The final global graph $\mathcal{G}=(\mathcal{N}, \mathcal{E})$ is synthesized by progressively merging the local subgraphs $\mathcal{G}_k=(\mathcal{N}_k, \mathcal{E}_k)$. We maintain $\mathcal{G}$, initialized with base nodes (e.g., Floor, Wall), and iteratively incorporate each $\mathcal{G}_k$. To process the edges from $\mathcal{G}_k$, we perform a Breadth-First Search (BFS) starting from edges connected to the base nodes in $\mathcal{G}_k$. For each edge $e_{new} = (o_i, r_{new}, o_j)$ in subgraph $\mathcal{G}_k$: If either object primitive $o_i$ or $o_j$ is not yet in the global node set $\mathcal{N}$ in $\mathcal{G}$, we add the new node and the edge $e_{new}$ directly to $\mathcal{G}$. However, if both $o_i, o_j \in \mathcal{N}$, we must check the new edge $e_{new} = (o_i, r_{new}, o_j)$ for potential conflict against the existing structure of $\mathcal{G}$. A conflict is identified if no path currently exists between $o_i$ and $o_j$, or if an existing path contains relationships inconsistent with $r_{new}$ or exhibits a disordered parent-child hierarchy. If such a conflict is detected, we initiate a conflict resolution: we identify all nodes $\mathcal{O}_{path}$ involved in the relevant path, re-optimize for an adjudication viewpoint $v_{adj}^*$ targeting $\mathcal{O}_{path}$, re-infer the relationship set $\mathcal{E}_{adj}$ among these nodes via VLM, and merge $\mathcal{E}_{adj}$ into $\mathcal{G}$, replacing existing wrong edges. Conversely, if a path exists and is consistent with $r_{new}$, we consider $e_{new}$ redundant and discard it, preserving the original graph structure. This iterative merging and conflict resolution process yields the final, globally consistent scene graph $\mathcal{G}$.

\paragraph{Hierarchical Physical Assembly.}
The synthesized scene graph $\mathcal{G}$ guides the following construction within the physical simulator. We initialize the environment by placing the base nodes Floor and Wall in $\mathcal{G}$ and designate them as passive rigid bodies. We then perform a Breadth-First Search (BFS) starting from these base nodes. For each new edge $e_{ij} = (o_i, r, o_j)$, where $o_i$ is the already placed parent object and $o_j$ is the child object to be placed. If the relation $r$ is support relationship, we place $o_j$ at its initial position $T_j$ but adjust slightly upwards relative to $o_i$. Object $o_j$ is momentarily set as an active rigid body, and physics simulation is briefly activated, allowing it to undergo realistic settling onto $o_i$'s surface via gravity and collision. Once settled, $o_j$ is typically converted back to a passive rigid body to ensure stability. Alternatively, if the relation $r$ is attachment relationship, we apply a fixed constraint between $o_i$ and $o_j$ to simulate anchoring $o_j$ directly onto $o_i$'s surface, mimicking a physical attachment. This hierarchical, physics-based assembly process, guided by the scene graph, ensures a natively plausible scene construction.
\section{Experiments}

\subsection{Experimental Settings}
\paragraph{Datasets.} We conduct experiments on 20 scenes from the real-world ScanNet dataset~\cite{dai2017scannet}, using only raw RGB videos as input, without access to depths, normals, or semantics.

\paragraph{Baselines.} For compositional scene reconstruction, we compare against state-of-the-art baselines DPRecon~\cite{ni2025dprecon} and InstaScene~\cite{yang2025instascene}. We also include comparisons with top-performing single-view methods, Gen3DSR~\cite{ardelean2025gen3dsr} and SceneGen~\cite{meng2025scenegen}, which take the target image as input. Additionally, to evaluate physical plausibility of the final simulation-ready scenes, we further compare against the 3D indoor simulator MetaScenes~\cite{yu2025metascenes}.

\paragraph{Metrics.} We assess our method using quantitative metrics for reconstruction and rendering. For reconstruction quality, we evaluate Chamfer Distance (CD), F-Score, and Normal Consistency (NC) following MonoSDF~\cite{yu2022monosdf}. For rendering fidelity, we adopt the full-reference (FR) and no-reference (NR) setup from ExtraNeRF~\cite{shih2024extranerf}. The FR metrics include PSNR, SSIM, and LPIPS, while for NR, we employ MUSIQ~\cite{ke2021musiq} to assess perceptual quality. Additionally, we report the average processing time of each method.

\paragraph{Implementation Details.} In this paper, we leverage 2DGS~\cite{huang20242d} for 3D reconstruction from video input, follow SceneSplat~\cite{li2025scenesplat} for semantic segmentation, perform single-object generation using Rodin~\cite{zhang2024clay} and finally construct the simulation-ready scenes in Blender and Issac Sim. Moreover, we adopt Qwen2.5-VL~\cite{bai2025qwen2} for intrinsic attributes inference and scene graph inference. We optimize our active viewpoint on a single NVIDIA RTX A6000 GPU with about 30 seconds for each object. More details of our framework are discussed in the supplementary material.

\begin{table*}[tp]
	\centering
    \caption{\textbf{Quantitative Comparison for Compositional 3D Reconstruction.} We evaluate our method against single-view (Gen3DSR~\cite{ardelean2025gen3dsr}, SceneGen~\cite{meng2025scenegen}) and scene-level (DPRecon~\cite{ni2025dprecon}, InstaScene~\cite{yang2025instascene}) baselines. The comparison includes metrics for geometric fidelity (CD, F-Score, NC), novel-view rendering quality (PSNR, SSIM, LPIPS, MUSIQ), and inference time.}
	\begin{tabular}{l|ccc|cccc|c}
		\toprule
		 \multirow{2}{*}{Method} & \multicolumn{3}{c}{Reconstruction} & \multicolumn{4}{c}{Rendering} & \multirow{2}{*}{Time}\\
         & CD$\downarrow$& F-Score$\uparrow$ & NC$\uparrow$  &PSNR$\uparrow$ &SSIM$\uparrow$ &LPIPS$\downarrow$ &MUSIQ$\uparrow$ \\
		\midrule
        
        Gen3DSR&11.69&30.19&70.50&19.26&0.886&0.425&60.94&\underline{17min} \\
        SceneGen&7.66&46.72&79.13&18.18&0.873&0.334&60.22&\textbf{6min} \\
        \midrule
        DPRecon&9.26&46.12&78.28&21.97&\underline{0.913}& \underline{0.257}&71.49&10h 42min \\
        InstaScene&\underline{6.90}&\underline{49.69}&\underline{82.55}&\underline{22.35}&0.907&0.302&\underline{71.57}&29min \\
        \textbf{Ours}&\textbf{4.34}&\textbf{62.65}&\textbf{87.37}&\textbf{24.43}&\textbf{0.924}&\textbf{0.153}&\textbf{73.56}&21min \\
		\bottomrule
	   \end{tabular}
    \label{main_table}
\end{table*}

\begin{figure*}[t]
    \centering
    \includegraphics[width=\linewidth]{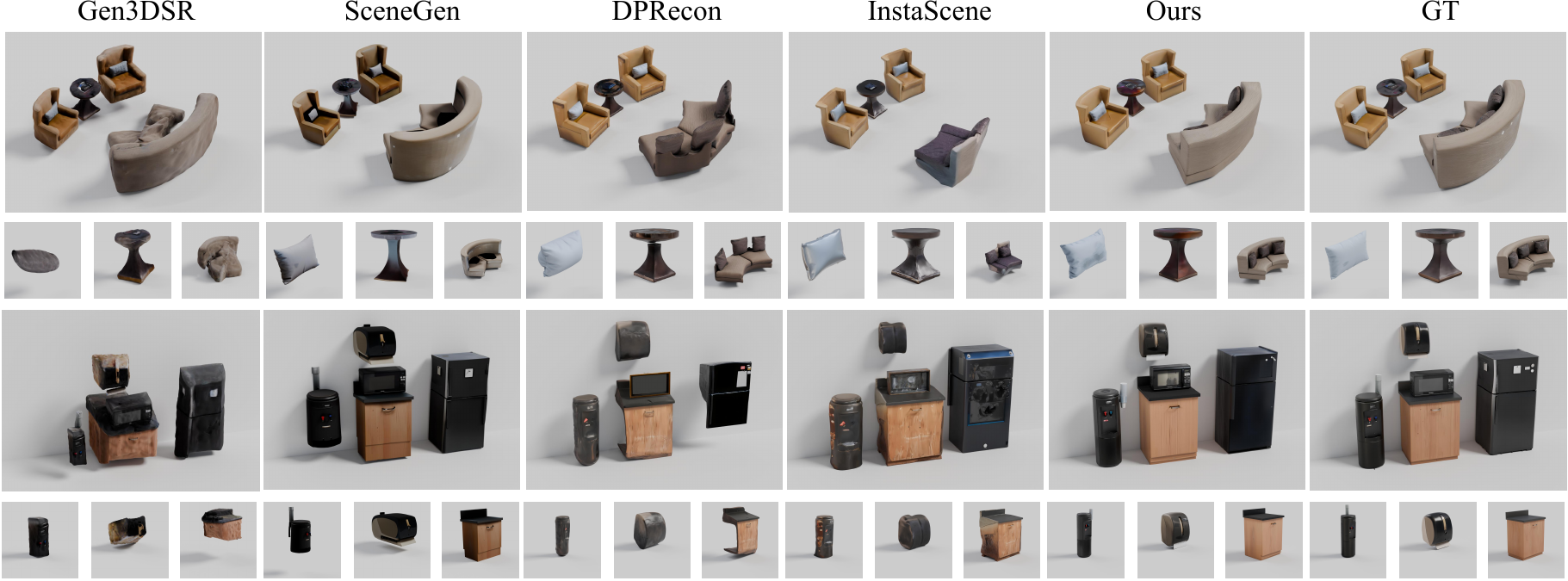}
    \caption{\textbf{Qualitative Comparison for Compositional 3D Reconstruction.} We present qualitative visualizations of the final reconstructed scenes. For single-view setting, we render the 3D representation at the target viewpoint as the input for these methods.}
    \label{main_vis}
\end{figure*}

 \begin{figure}[t]
    \centering
    \includegraphics[width=\linewidth]{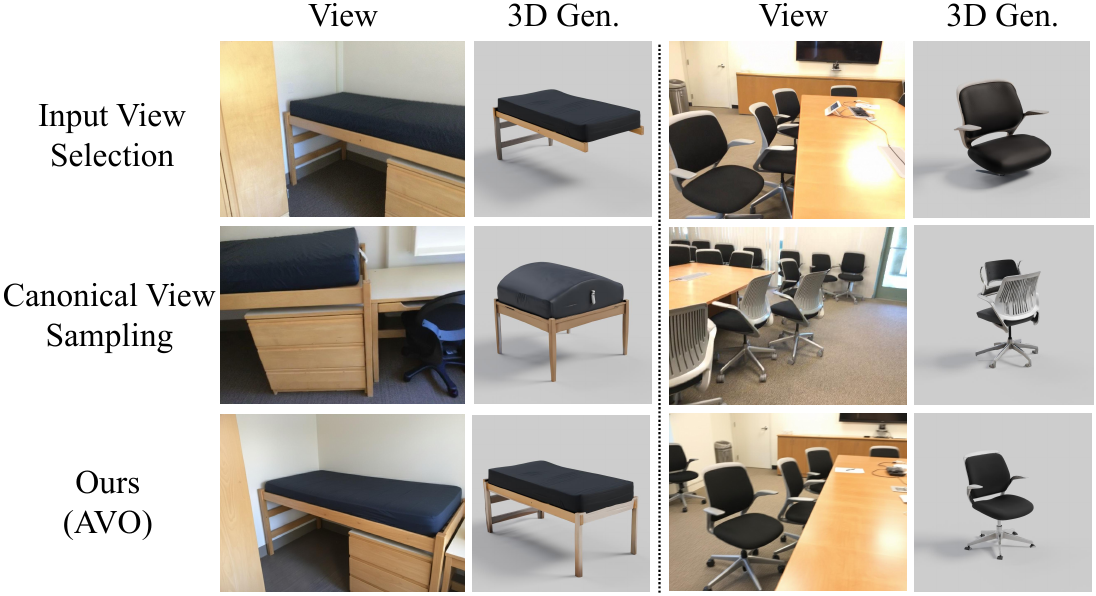}
    \caption{\textbf{Qualitative comparison of viewpoint sampling strategies.} We uniformly use a single image as the condition and utilize the same generative model.}
    \label{fig:comparison_avo}
    \vspace{-6mm}
\end{figure}

 \begin{figure}[t]
    \centering
    \includegraphics[width=\linewidth]{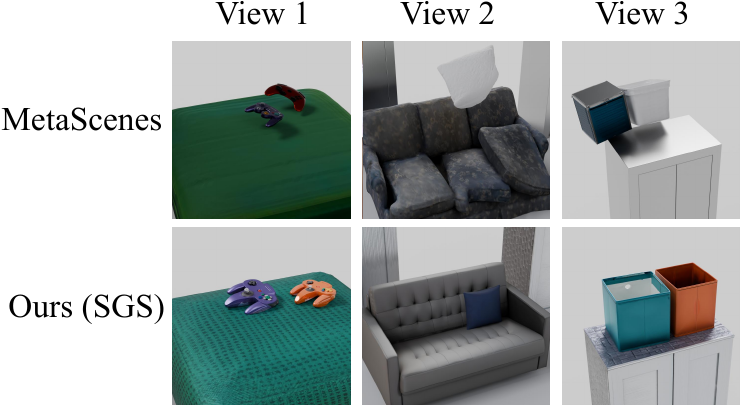}
    \caption{\textbf{Qualitative comparison of physical scene construction in the simulator.}}
    \label{fig:comparison_sgs}
    \vspace{-6mm}
\end{figure}

 \begin{figure}[t]
    \centering
    \includegraphics[width=\linewidth]{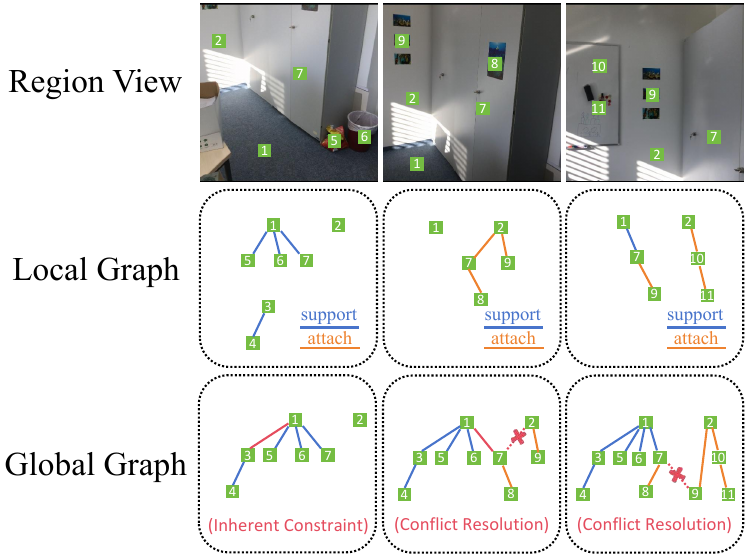}
    \caption{\textbf{Visualization of the progressive Scene Graph Synthesizer.} Optimal region images are captured (top row), from which local current graphs are inferred (middle row) and then progressively merged into the final global graph (bottom row). We use green signals for instance IDs and red signals for special edges.}
    \label{fig:visualization_sgs}
    \vspace{-5mm}
\end{figure}

\subsection{Results}
\paragraph{Compositional 3D Reconstruction.} Table~\ref{main_table} and Figure~\ref{main_vis} present the quantitative and qualitative results for the compositional 3D reconstruction task. We observe that single-view methods like Gen3DSR~\cite{ardelean2025gen3dsr} and SceneGen~\cite{meng2025scenegen} struggle to reconstruct faithful object geometry with accurate spatial positions, demonstrating limited generalization ability to real images. DPRecon~\cite{ni2025dprecon}, which employs a signed distance field (SDF) of each object as a strong 3D generative condition, consequently suffers from deformed artifacts stemming from the heavily incomplete 3D structure, which also costs significant inference time. InstaScene~\cite{yang2025instascene}, which leverages a heuristic view sampling strategy on semantic 3D Gaussians as conditions, often yields heavily occluded projected images, consequently failing to generate accurate geometry and appearance. In contrast, our method employs Active Viewpoint Optimization to intelligently search for optimal projections by maximizing 3D information gain, facilitating the reconstruction of assets with high geometric and visual fidelity. Moreover, our framework utilizes the synthesized scene graph to dictate the physics-based asset assembly, ensuring a physically plausible final configuration without floating or penetrated situations.

\paragraph{Projected Images Comparison.} Figure~\ref{fig:comparison_avo} presents the visualization results of three distinct viewpoint sampling methods. First, we evaluate the input view with maximum 2D object visibility, but this sampling objective is often insufficient to guide the complete object generation, due to the discrepancy between its 2D pixel coverage and the required 3D structural information. Second, we sample canonical views around the target object, but this strategy still yields occluded perspectives, resulting in malformed geometry and appearance. In contrast, our method actively optimizes for an ideal viewpoint in 3D space capturing the full structure and appearance of the target object, proving successful as the condition for 3D generation models, and robustly adaptive to target objects of varying scales.

\paragraph{Physical Construction Comparison.} Figure~\ref{fig:comparison_sgs} presents the visualization results in the 3D simulator Blender against MetaScenes~\cite{yu2025metascenes}, which provides 3D indoor simulation data derived from ScanNet~\cite{dai2017scannet}. This method relies on well-reconstructed 3D point clouds as input and primarily employs a retrieval-based strategy to acquire objects, which results in a lack of fidelity to the original scene. For the final physical scene construction, MetaScenes relies on a post-hoc Markov Chain Monte Carlo (MCMC) search to merely resolve collisions, an inefficient ``blind" optimization that is prone to local optima and fails to model accurate contact relationships. In contrast, our framework adopts a physically native approach, leveraging the synthesized scene graph to guide a hierarchical, physics-informed assembly that natively ensures both semantic coherence and physical stability from the outset. We further provide illustrative visualizations for our scene graph synthesizer in Figure~\ref{fig:visualization_sgs}.

\subsection{Ablation Study} 
\begin{figure}[t]
    \centering
    \includegraphics[width=\linewidth]{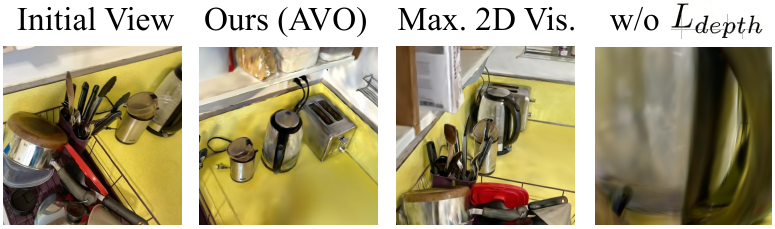}
    \caption{\textbf{Ablation study for our Active Viewpoint Optimization (AVO).} Here, the target object is the kettle and Max. 2D Vis. denotes the baseline using the optimization objective of maximizing 2D object visibility.}
    \label{fig:ablation_avo}
    \vspace{-2mm}
\end{figure}

\begin{figure}[t]
    \centering
    \includegraphics[width=\linewidth]{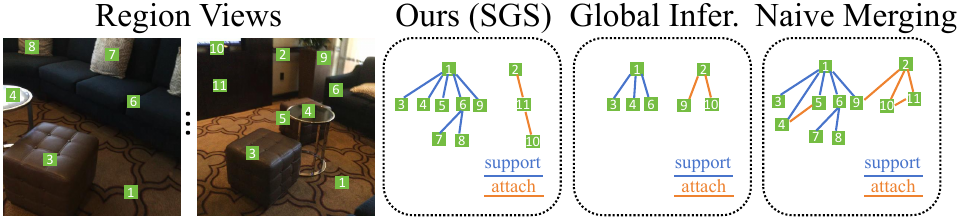}
    \caption{\textbf{Ablation study for our Scene Graph Synthesizer (SGS).} Here, green signals represent the instance IDs. Global Infer. and Naive Merging represent the baseline with a single global inference and the baseline that simply merges subgraphs without conflict resolution respectively.}
    \label{fig:ablation_sgs}
    \vspace{-6mm}
\end{figure}

We conduct extensive ablation studies to validate the effectiveness of our two bridging modules. For the Active Viewpoint Optimization (AVO) module, we visualize the projected images from two ablated settings in Figure~\ref{fig:ablation_avo}: an optimization result based solely on maximum 2D visibility, and our optimized result without the $L_{depth}$ supervision. The maximum 2D visibility baseline often just covers the whole object without further refinement, while omitting the depth constraint results in viewpoints that collapse impractically close to the object surface. For the Scene Graph Synthesizer (SGS) module, we visualize the synthesized scene graphs from two ablated strategies in Figure~\ref{fig:ablation_sgs}: a single inference on a global projected image, and a naive merging process lacking conflict resolution. The single inference approach fails to capture all objects and their relations accurately, and the naive merging strategy produces an incoherent graph with messy relationships, unsuitable for guiding the subsequent construction process.

\section{Conclusion}

In this paper, we propose SimRecon, a ``Perception-Generation-Simulation" pipeline designed to create object-centric, simulation-ready scenes from cluttered real-world videos. Our framework addresses the critical stage transition barriers that cause visual infidelity and physical implausibility in naive pipeline combinations. We introduce two key bridging modules: Active Viewpoint Optimization, which actively searches for optimal projections to ensure high-fidelity generative conditions, and a Scene Graph Synthesizer, which guides a constructive assembly that mirrors the real construction principle to ensure physical plausibility from the outset. Experiments on the ScanNet dataset validate that our method achieves superior performance in both reconstruction quality and physical adherence.

{
    \small
    \bibliographystyle{ieeenat_fullname}
    \bibliography{main}
}


\end{document}